# Tree level change detection over Ahmedabad city using very high resolution satellite images and Deep Learning

Jai G Singla*  
Space Applications Centre (ISRO)  
jaisingla@sac.isro.gov.in

Gautam Jaiswal  
HMR institute of Technology and Management, Delhi  
gautamjaiswal030@gmail.com

# Tree level change detection over Ahmedabad city using very high resolution satellite images and Deep Learning

**Abstract:** In this study, 0.5m high resolution satellite datasets over Indian urban region was used to demonstrate the applicability of deep learning models over Ahmedabad, India. Here, YOLOv7 instance segmentation model was trained on well curated trees canopy dataset (6500 images) in order to carry out the change detection. During training, evaluation metrics such as bounding box regression and mask regression loss, mean average precision (mAP) and stochastic gradient descent algorithm were used for evaluating and optimizing the performance of model. After the 500 epochs, the mAP of 0.715 and 0.699 for individual tree detection and tree canopy mask segmentation were obtained. However, by further tuning hyper parameters of the model, maximum accuracy of 80 % of trees detection with false segmentation rate of 2% on data was obtained.

## I. INTRODUCTION

Vegetation has very important role of eliminating harmful pollutants, reducing noise, regulating the environment's temperature, conserving water sources, and providing oxygen. Currently, governments across the globe are focusing to tackle environmental challenges and reduce carbon pollution which incorporates the latest technology such as Artificial Intelligence (AI) and deep learning to develop solutions to overcome these challenges. Detection, segmentation, and counting of individual trees using remote sensing and deep Learning helps to develop applications for urban planning, monitoring of tree vegetation, water preservation in urban areas, climate change, etc. In this context, Fei Tong et al (2021) implemented the Gaussian binarization with spatial local maxima for automatic delineation and detection of individual tree crown borders and located treetops on various tree species located in the district of Black Brook, New Brunswick, Canada[1]. They achieved the accuracy between 48% to 75%. Abdellatif Moussaid et al (2021) proposed an improved approach for the segmentation of tree crowns and classification of overlapping trees in orchards. They have implemented the K-means clustering algorithm on 0.5 m spatial resolution satellite imagery of citrus trees located in Ouargha Orchard, Morocco. They have classified the segments of overlapped citrus trees crowns and estimated their sizes with F1-score of 0.93 with the precision of 0.93[2]. Rebeka Bhattacharyya et al (2019) implemented the CNN-based LeNetCNN model for the detection and counting of shade trees using satellite images. In this research, they have implemented the model on satellite images over the two different regions of Jalpaiguri district, India. The overall accuracy of the model was greater than 95% for correctly detecting the shade trees and less than 4% error to count the number of shade trees[3]. Weijia Li et al (2019) proposed real-time tree crown detection on large-scale remote sensing satellite raw images on a Field Programmable Gate Arrays (FPGAs). In this research, they have designed a pipelined-friendly tree crown detection approach (PF-TCD) and implemented it on papaya and oil palm tree aerial LiDAR datasets. This algorithm obtained the F1 scores between 82% to 92% 18 times faster than other methods for real-time tree detection on a large scale of 12,188 × 12,576 pixels remote sensing images in 0.33s[4].

Ben G. Weinstein et al (2019) used the semi-supervised deep learning model RetinaNet for detecting and classifying the tree's crown over RGB satellite imagery. In this paper, they have used the NEON 2018 dataset over the San Joaquin, CA, and implemented the comparative unsupervised, self-supervised, and full semi-supervised model in which the semi-supervised model had an average tree crown recall of 0.69 and precision of 0.61 with the average tree detection rate of 82% over the field[5]. Weijia Li et al (2018) proposed a two-stage Convolutional neural network based on different architectures such as AlexNet, LeNet, and VGG-19 with the support vector machines

(SVM), random forest (RF), and ANN followed by land classification and tree object classification. They have used the QuickBird panchromatic and multi-spectral remote sensing data over the south of Malaysia and evaluated the detection results of all methods in terms of F1 score in which the combination of AlexNet and LeNet achieved the highest average F1 score of 95% among all 11 methods[6]. Milad Mahour et al (2016) worked on Gaussian scale-space theory for the automatic detection of individual trees from very high-resolution satellite Images with different sizes, species, spatial variation, and limited spatial information. In this paper, they optimized the Gaussian algorithm for the detection and extraction of trees in two orchards both located in the Qazvin agricultural area in North-western Iran, and implemented on WV-2 VHR satellite image and an UltraCam digital aerial imagery dataset of trees. This research showed 95% detection accuracy with a greater than 0.29 false error for counting the trees[7]. Weijia Li et al (2017) also worked on oil palm tree detection over Malaysia using convolutional neural network. They could detect more than 96% of palm trees correctly as compared with ground survey[8]. John Secord et al (2006) proposed a two-step method for tree detection with segmentation followed by classification on aerial images obtained by LiDAR. In this paper, they have used the Airborne LiDAR dataset of Berkeley, California, including residential, commercial, and University of California, Berkeley campus areas. For segmentation, they used a spectral clustering algorithm using learning weight and to classify the segments of trees they used a support vector machine (SVM) had achieved a ROC score of overall 97%[9].

The objective of our work is to illustrate the potential of deep learning-based instance segmentation methods for estimating the change detection and counting individual trees from the tree instances identified in high-resolution remote sensing satellite images (0.5m spatial resolution). We generated curated datasets for training based on input satellite images, then trained the "state-of-the-art" deep learning model to detect and segment the individual tree instances. Then, we implemented the trained model to count the number of individual tree and performed the change detection on time series data of 2011 and 2018 satellite data over Ahmedabad region.

## II. RELATED WORKS

Recently, many researcher worked for detection of individual trees and change detection using high-resolution satellite images using deep learning. Chen Qian et al (2023) derived the methodology based on 3-dimensional geometric features of tree crowns from Airborne LiDAR cloud to segment the individual trees. In this paper, they have implemented the local maximum algorithms and improved rotating profile-based delineations method through the tree shaping of the Digital Terrain Model (DTM) and Digital Surface Model (DSM) for Crown Height Model (CHM) generation and tested over the two different sites located at Tiger and Leopard National Park in Northeast China. The study results indicated an average tree classification accuracy of 90.9% and the optimal classification accuracy reached 95.9% on field [10]. Pengcheng Wang et al (2023) combined the semantic and instance segmentation model for automatic extraction and instance-level segmentation of roadside trees from vehicle-mounted mobile laser scanning (MLS) point clouds data. They implemented the segmentation approach based on graph semantic and feature encoders-decoders and feature fusion for segmenting the crowns of the individual trees in urban areas. This novel method accurately segmented approximately 90% of roadside trees with a precision of 88% and recall of 87% [11]. Lihong Chang et al (2022) proposed state of an art two-stage semantic and instance segmentation approach for the segmentation of individual trees from the TLS point clouds dataset from the international TLS benchmarking project. They used the RandLA-Net with SGD optimizer to train the model over the dataset for semantic segmentation and then implemented the instance segmentation approach using YOLOv3 through transfer learning to detect and classify the individual instance of trees. The state-of-the-art model obtained a mean accuracy of 90 % with correctness of 95% on the TLS benchmark dataset [12]. Yingbo Li et al (2022) implemented the Mask R-CNN (Mask region-based convolution neural network) with the proposed edge detection R-CNN (ACE R-CNN) based instance segmentation and attention complementary network (ACNet) for individual tree species segmentation in high-density forests on RGB and LiDAR data. They implemented the proposed algorithm on Gaofeng Forest Farm which is a forest plantation in Nanning, Guangxi Province, southern China. The obtained results of ACE R-CNN in terms of precision, recall, F1-score, and Average Precision were 0.97, 0.94, 0.95, and 0.96 respectively[13]. Qiuji Chen et al (2021) combined the DBSCAN and K-Means clustering algorithm for improving the single tree segmentation algorithm for segmenting the single tree. In this paper, the research work is implemented on airborne LiDAR point cloud data of an underground coal mine in Shenmu City, Shaanxi Province, China. This improved method shows the segmentation accuracy in terms of F1 score is 0.894 and 0.966 for leaf and leafless trees segmentation[14].

Anastasiia Safonova et al (2021) used the Mask R-CNN instance segmentation for olive tree crown and shadow segmentation (OTCS) that further help to estimate the biovolume of individual trees. They built their own multi-spectral orthoimages UAV dataset of olive tree crowns and implemented the Mask R-CNN model on the dataset for segmentation of trees and shadow of trees. The model obtained an F1-score greater than 96% with a minimum false negative error of 5.4%[15]. Lino Comesana-Cebral et al (2021) implemented the segmentation of individual trees from raw LiDAR point clouds obtained by a backpack TLS in the Spanish region of O Xures, Galicia, Spain. They improved the method based on DBSCAN clustering and cylinder voxelization that achieved detection rates near 90% and low commission and submission errors accuracy of over 93%[16]. Nils Nolke (2021) presented the neural network-based approach for detecting and mapping the trees. They implemented the Multi-layer Perceptron (MLP) deep learning network on 30 cm remote sensing data from WorldView-3 and Landsat 8 from November 2016 over the Northern parts of the Bangalore Metropolitan Region to map the tree cover areas and compared with Global 30 m Landsat Tree Canopy Version 4 that results in the mean absolute error (MAE) = 13.04%[17]. Xingcheng Lu et al (2014) developed the bottom-up method based on the intensity and 3D structure of leaf-off LiDAR point cloud data for tree segmentation in deciduous at the Shavers Creek Watershed in Pennsylvania. In the bottom-up method, they first extracted and segmented the tree trunks through the intensity and topological allocated points and then allocated the other points less than 35 thresholds to extract the tree trunks on 2D and 3D distances. The bottom-up algorithm obtained the F1-score of 90%, recall of 84%, and precision of 97% on leaf-off LiDAR point cloud data [18]. Linhai Jing et al (2012) proposed an innovative multi-scale analysis and segmentation (MSAS) method with different Gaussian settings to segment the tree crowns over the LiDAR data. This study work was implemented near the forest site of Sault Ste. Marie, Ontario, Canada, within the Great Lakes-St. Lawrence forest region. The overall accuracy statistics of the MSAS method to generate a tree crown map was 73% with omission and commission errors of 23 and 10% respectively[19]. George Chen et al (2009) presented a two-stage approach based on segmentation followed by classification of trees in large urban areas using Aerial LiDAR data. In this paper, they implemented the Random Forest Classifier on a combined dataset of the North American dataset and a European dataset of Trees. The resulting accuracy of the implementation in terms of precision and recall is over 95% [20].

Most of the studies described above are carried out on aerial data using LiDAR and UAVs for tree detections. A few studies also used satellite data for the tree detection purpose but not exploited it to full extent. In this study, we have worked on 0.5m high resolution satellite datasets over Indian region. Our study demonstrates the applicability of deep learning latest models over Ahmedabad, India. As, city of Ahmedabad contains dense tree cover; it becomes very challenging to detect and count individual trees from satellite data. However, we made use of "state-of-the-art" YOLOv7 instance segmentation algorithms for tree detection, individual tree count, and automatic change detection over tree canopy. We could obtain results with 71% accuracies.

### III. METHODOLOGY

As a part of methodology, remote sensing satellite data of 50 cm of two different times are obtained to achieve the objective. A fair amount of tiles were generated using the data and these tiles are further labelled with tree level details. Further, these tiles are split into ratio of training , testing and validation tiles in order to evaluate the performance of the model. Further, trained model was used to predict change in tree canopy for two different times. Entire methodology is described as Figure-1 and model architecture is stated in Figure-2.

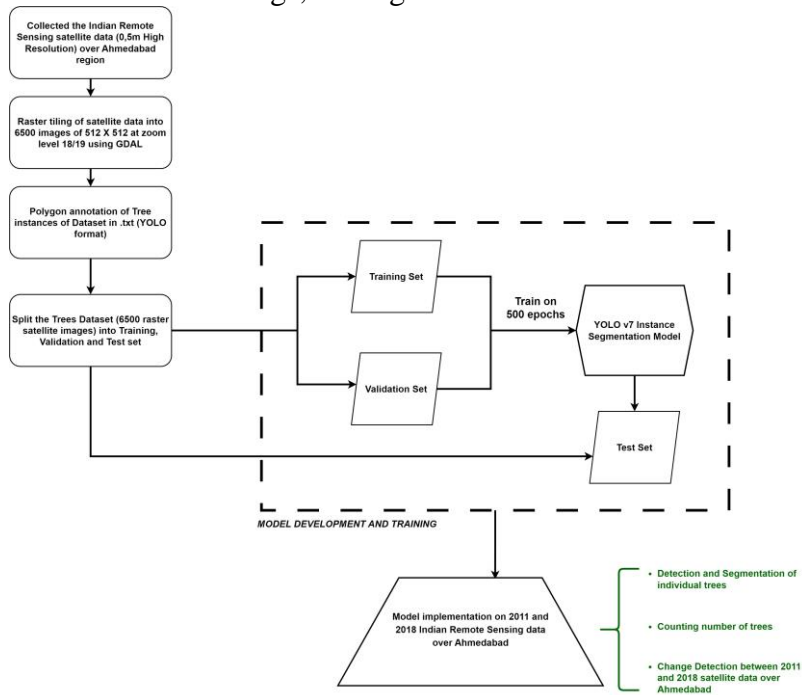

*Figure 1: The workflow of Methodology.*

**(a) Dataset Details**

There is a requirement of high-resolution satellite imagery to proceed for accurate detection and counting of tree over a site. We made use of very high-resolution (50cm image data) of World View (WV) -2 satellite as an input for this task. WV-2 is high-resolution commercial satellite launched by Digital Globe in the United States. WV-2 provides panchromatic imagery with resolution of 0.46m and multispectral resolution of 1.85m. Brief description of the data is mentioned in Table-1.

| Images | WorldView-2 Image 1 | WorldView-2 Image 2 |
|---|---|---|
| Acquisition Date (D/M/Y) | 2011-01-25 | 2018-12-7 |
| Acquisition angle | 17 degree approx. | 22 degree approx. |
| Ground resolution | 50cm | 50cm |

| Map Projection | UTM Zone43 North WGS84 | UTM Zone43 North WGS84 |
| --- | --- | --- |
| Bands | 4 Bands | 4 Bands |
| Total Area | 61 sq km | 61 sq km |
| xMin,yMin | 72.4587,22.9942 | 72.4587,22.9942 |
| xMax,yMax | 72.5716,23.0835 | 72.5716,23.0835 |
| Image quality | 16 bit | 16 bit |
| Registration accuracies | Better than 1 pixel | Better than 1 pixel |

Table -1 : Detailed data description

**(b) Dataset Preparation and Preprocessing**

For the study, we have used the 0.5m high-resolution satellite imagery over the Ahmedabad region. During the preparation of the dataset, we made tiles of satellite geospatial data into 6500 raster images of 512x512 pixels at the zoom level of 18/19 using GDAL (Geospatial Data Abstraction Library) package. Further, LabelMe image annotation software tool was used to perform polygon annotation to label the tree instances from the images. Further, data (images and annotations) was split into 70% for training, 20% for validation and 10 % for validation on unseen data.

**(c) Model Architecture**

Wang et al. (2022) introduced YOLOv7 as an improved version of YOLOv5, specifically designed for image segmentation tasks, with a focus on instance segmentation. YOLOv7 showcased significant advancements in both speed and accuracy, surpassing previous real-time object detection and instance segmentation algorithms.

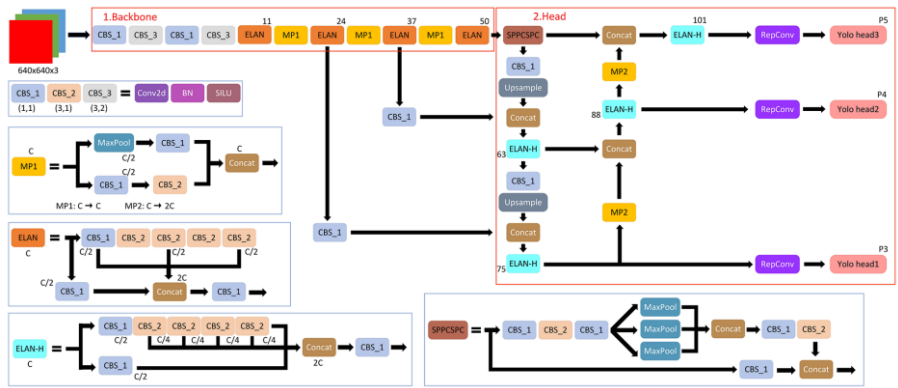

*Figure 2: The architecture of YOLO v7.*

Due to all the advantages of YOLO V7, we trained this model on curated datasets.

**(d) Evaluation Metrics and Loss Function**

To evaluate the results, we have used the precision, recall, and mean average precision of both detection and segmentation at Intersection Over Union (IoU) thresholds of 0.5 and 0.95 to measure the accuracy of predicted bounding boxes and segmentation masks. Precision measures the percentage of correctly predicted results in which true positives represent the correctly predicted instances (bounding boxes or segmentation masks) with a very high overlap with the ground truth instances and False positives represent the incorrectly predicted instances. Recall measures the percentage of correctly predicted results at IoU 0.5:0.95 or higher in which True positives represent the correctly predicted instances (bounding boxes or segmentation masks) that have an IoU of 0.5 or higher with the ground truth instances. False negatives represent the instances that were not detected or missed by the model.

$$Precision\ (P) = \frac{True\ Positive}{True\ Positive + False\ Positive} \quad (1)$$

$$Recall\ (R) = \frac{True\ Positive}{True\ Positive + False\ Negative} \quad (2)$$

Mean Average Precision (mAP) calculates the mean value of the average area under the precision-recall curve values for all classes by evaluating the different confidence thresholds. mAP is considered an important evaluation metric to measure the overall performance of a model in which higher mAP indicates better instance segmentation and object detection performance with values lying between 0 to 1.

$$Mean\ Average\ Precision\ (mAP) = \frac{1}{n}\sum_{1}^{n} Average\ Precision_k \quad (3)$$
$$n = the\ number\ of\ classes$$
$$k = particular\ class$$

During the training of the model, we used two loss functions Bounding box regression loss and Mask regression loss to measure the performance of the targeting task and quantify the dissimilarity between the prediction results and ground truth annotation.

$$L_{box} = \lambda_{coord} \sum_{i=0}^{S^2} \sum_{j=0}^{B} [obj_{ij}\ (\lambda_x\ (t_x - b_x)^2 + \lambda_y(t_y - b_y)^2 + \lambda_w(t_w - b_w)^2 + \lambda_h(t_h - b_h)^2)] \quad (4)$$

$L_{box}$ is the Bounding Box Regression Loss is calculated using smooth L1 loss between the predicted box coordinates (x, y, width, height) and the ground truth box coordinates. In which, $\lambda_{coord}$ is a coefficient to balance the impact of the bounding box regression loss compared to other loss terms. $S$ is the grid size. $B$ is the number of anchor boxes per grid cell. $obj_{ij}$ is an indicator function that evaluates to 1 if the $i$th grid cell of the $j$th anchor box is responsible for detecting an object, and 0 otherwise. $t_x, t_y$ are the predicted center coordinates of the bounding box in the $i$th grid cell of the $j$th anchor box. $b_x, b_y$ are the ground truth center coordinates of the bounding box in the $i$th grid cell of the $j$th anchor box. $t_w, t_h$ are the predicted width and height of the bounding box in the $i$th grid cell of the $j$th anchor box. $b_w, b_h$ are the ground truth width and height of the bounding box in the $i$th grid cell of the $j$th anchor box. $\lambda_x, \lambda_y, \lambda_w, \lambda_h$ are coefficients to balance the impact of each coordinate term in the bounding box regression loss.

Mask Regression Loss which calculates the pixel-wise binary cross-entropy loss to compare the predicted pixel-wise mask probabilities with the ground truth masks.

$$L_{BCE} = -\frac{1}{N}\sum_{i=1}^{N}(y_i\ \log(p_i) + (1 - y_i)\log(1 - p_i)) \quad (5)$$

where $L_{BCE}$ is the pixel-wise binary cross-entropy loss, $N$ is the number of pixels, $y_i$ is the ground truth label for pixel instance $i$, and $p_i$ is the predicted probability for pixel instance $i$ belonging to the generated mask.

**(e) Hyperparameters**
Table 2 enlists some important Hyperparameters and alternate values, which were used during the model's training. The important hyperparameters to train instance segmentation models are number of soft anchor boxes per image, mask ratio, batch size and learning rate, in which soft anchor boxes define the number of auto-generated bounding boxes to detect the instances of objects and mask ratio specifies the segmentation of object region during the training of the model. The learning rate specifies the extent of gradient values toward a minimum of a loss function at each iteration. We have used Stochastic Gradient Descent (SGD) optimizer with the batch size of 16 and 32, 0.938 momentum, weight decay between 0.0005 to 0001 to effectively faster the performance and optimize the weights with the model complexity.

| Hyperparameters | Values |
|---|---|
| Learning rate (lr) | {0.01, 0.03} |
| Epochs | 500 |
| Batch size | {16, 32} |
| Momentum | 0.938 |
| Weight decay | {0.0005, 0.001} |
| Optimizer | SGD |
| Mask ratio | 0.4 |
| Anchor boxes per image | {4, 8} |

*Table 2. Hyperparameters during model training.*

**(f) Optimization Algorithm**
For improving the performance of our model, we have used the SGD optimizer algorithm with the hyperparameters such as weight decay, learning rate, and momentum. Stochastic Gradient Descent (SGD) is an advanced variant of the gradient descent algorithm that improves the convergence speed and performance of the model by updating the parameters continuously based on the computed gradient loss on small batches of training data.

$$v_{t+1} = \mu \cdot v_t - \eta \cdot \nabla J(\theta_t) - \lambda \cdot \eta \cdot p_t \quad (6)$$

$$p_{t+1} = p_t + v_{t+1} \quad (7)$$

$v_t$ is the momentum at time step $t$.
$\mu$ is the momentum coefficient.
$\eta$ is the learning rate.
$\nabla J(\theta_t)$ is the gradient of the loss function $J$ with respect to the parameters $\theta_t$.
$\lambda$ is the weight decay coefficient.
$p_{t+1}$ is the updated parameter at time step $t + 1$.
$\theta_t$ is the current parameter at time step $t$.

## IV. Results

All the experiments during the study are supported by the Personal Computer (PC) with the x86 64-bit Linux 7.7 (Mapio) operating system. The software configuration consists of Python (version 3.9) programming language, PyTorch framework 1.11, CUDA Toolkit 12.0, and Nvidia cuDNN 8.81. The hardware capabilities include NVIDIA Quadro GV100 GPU (64GB memory), Intel Xeon Platinum 8180 CPU @2.50GHz, and 1TB RAM. Figure 3,4 and 5 showcased the accuracy of the model in terms of recall of 0.70 , mAP > 0.7 an regression values ~0.7.

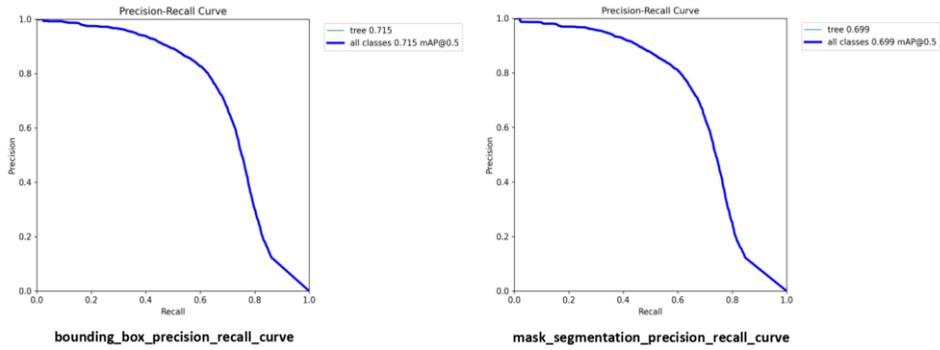

*Figure :3 The Box and Mask Precision and Recall Curve.*

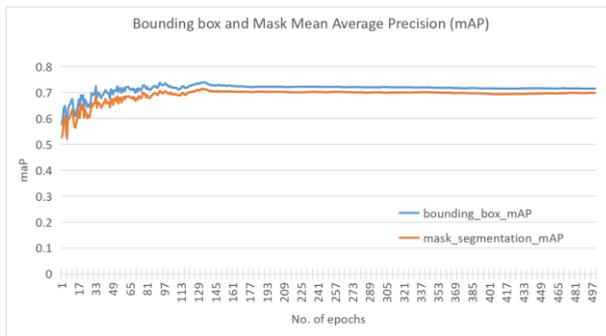

*Figure :4 The Mean Average Precision (mAP) of bounding box and mask segmentation.*

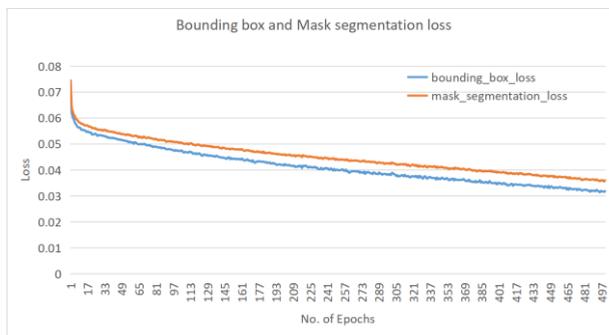

*Figure :5 The Bounding box regression and Mask Regression loss.*

Apart from results of evaluation metrics, Tree change detection results are generated over five different regions over Ahmedabad (Refer Figure:6). Results over each of the region are stated in Figure 7,8,9,10 and 11. It is evident from the Figures that trained model has performed effectively on given datasets and produced the change detection results at individual tree level and tree canopy in dense urban region.

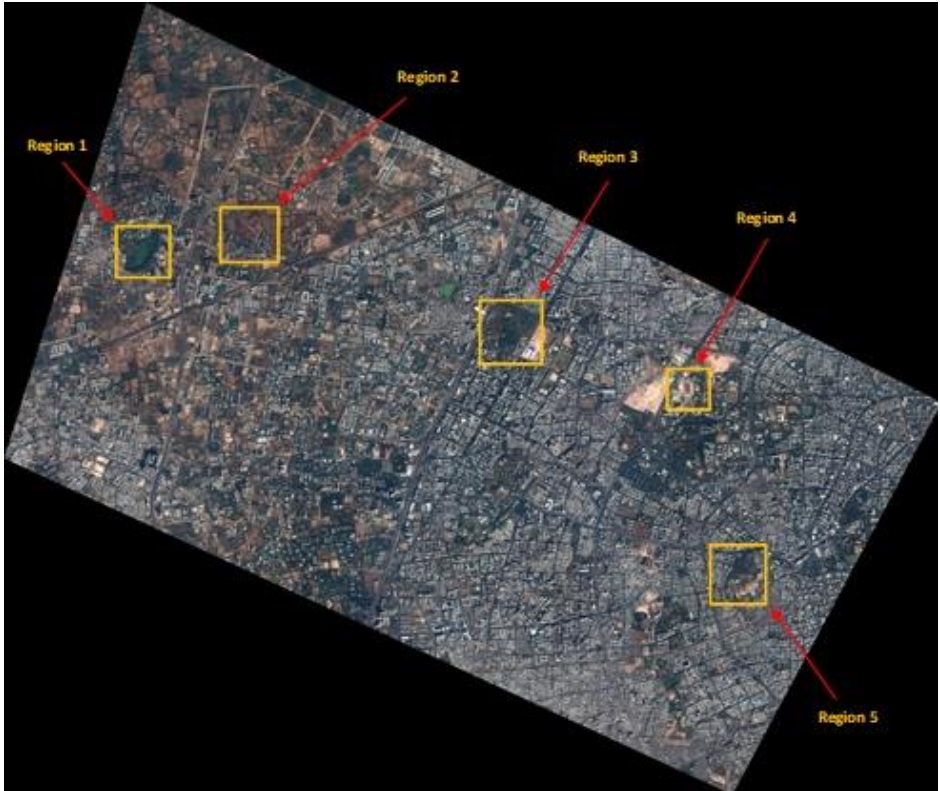

*Figure:6 Satellite image of Ahmedabad region of 2018. Visible 5 regions that represents the major changes from 2011 to 2018.*

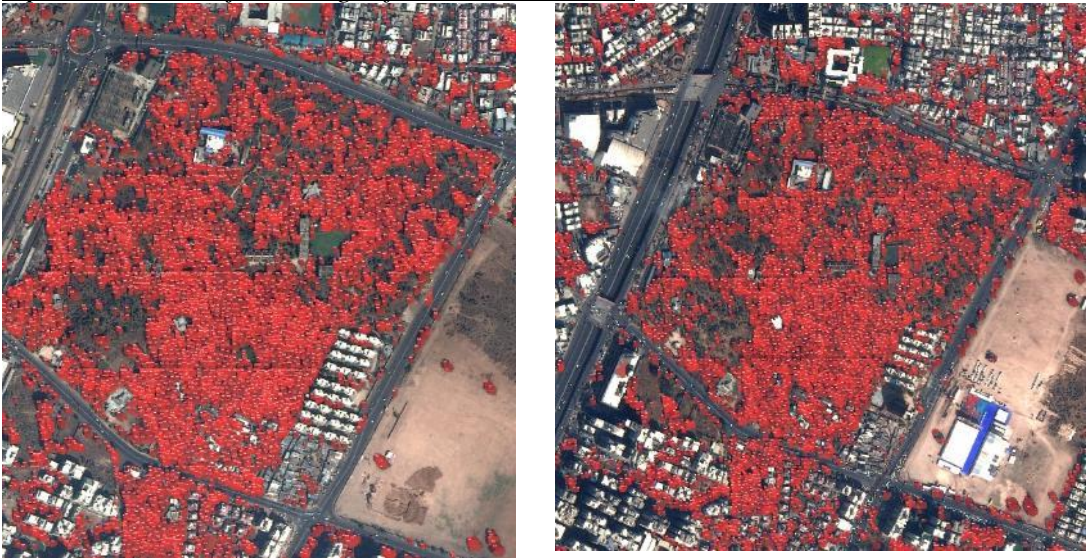

*Figure:7 Change Detection over region 3 of Ahmedabad from 2011 to 2018. Red mask represents the detected trees.*

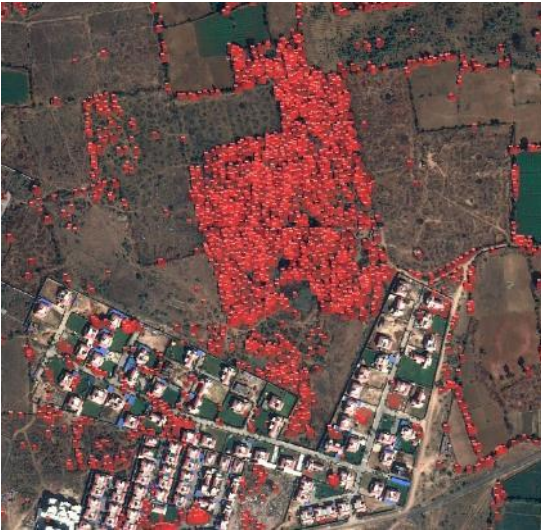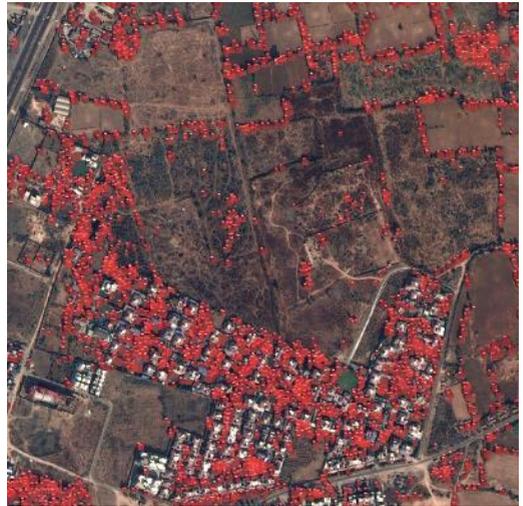

*Figure:8 Change Detection over region 2 of Ahmedabad from 2011 to 2018. Red mask represents the detected trees.*

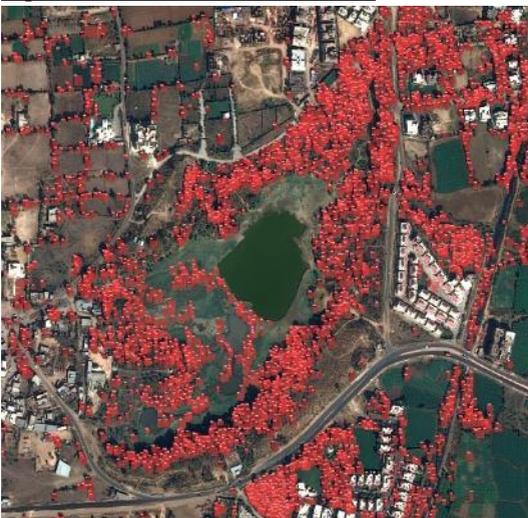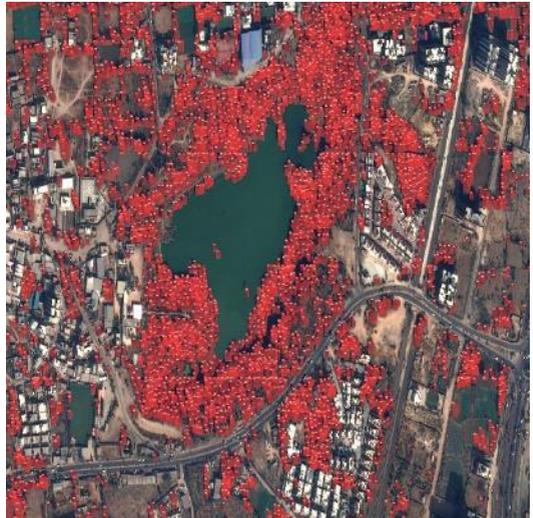

*Figure:9 Change Detection over region 1 of Ahmedabad from 2011 to 2018. Red mask represents the detected trees.*

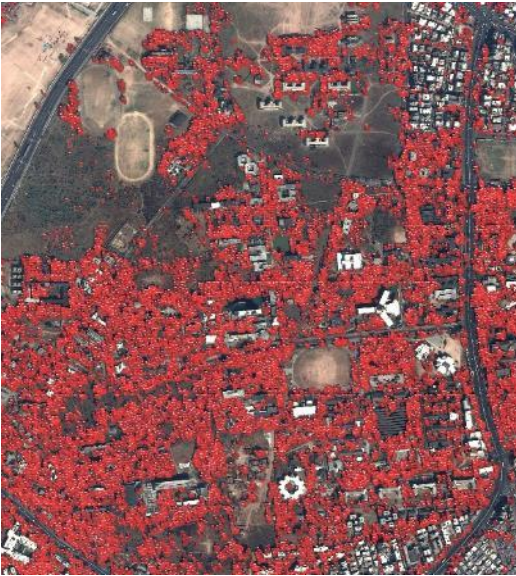 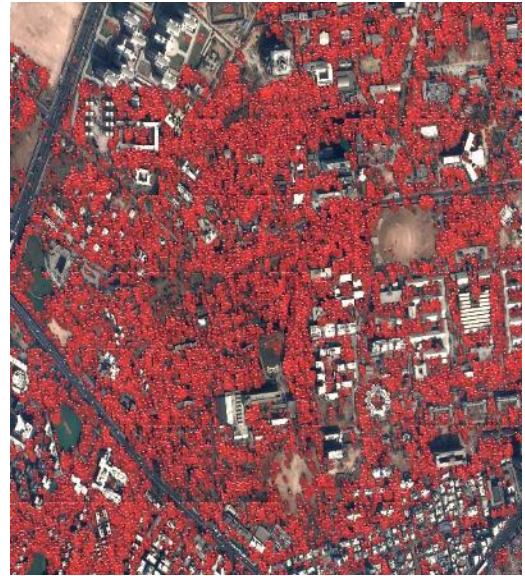

*Figure:10 Change Detection over region 4 of Ahmedabad from 2011 to 2018. Red mask represents the detected trees.*

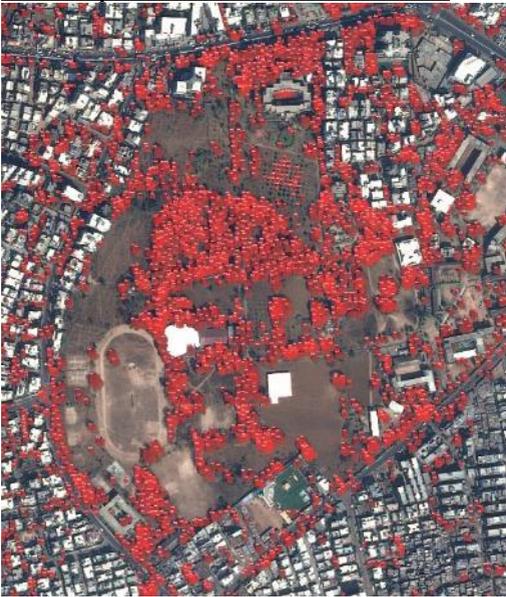 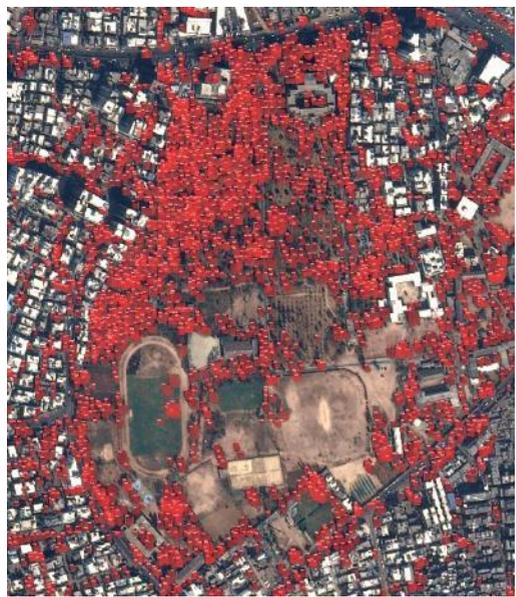

*Figure:11 Change Detection over region 5 of Ahmedabad from 2011 to 2018. Red mask represents the detected trees.*

## V. DISCUSSIONS

In this study, the instance segmentation model and 50cm high resolution remote sensing satellite data was trained. Deep learning model performed instance segmentation, counted the number of trees, and automatic change detection of tree canopy was carried out. Trained model correctly segmented and count the trees with the accuracy of 71% and 2% of false segmentation over the 2011 and 2018 satellite data of Ahmedabad which is relatively better than other proposed models for the related problems. After training the model on 500 epochs, results of 72% bounding box and 69% mask mean average precision (mAP) with the 0.072 segmentation loss were obtained. After evaluating model performance, it was concluded that the model used in this paper was not able to further optimize the segmentation loss and improve the mean average precision. The major challenge was to segment individual trees using 50 cm data, which often requires data of 5cm to 15cm resolution. However, by tuning hyperparameters of the model, maximum accuracy of 80 % of trees detection with false segmentation rate of 2% on data was obtained. In urban areas, the performance of the model is comparatively lower as compared to rural areas of Ahmedabad.

## VI. CONCLUSION AND FUTURE SCOPE

In this paper, the "state-of-the-Art" YOLOv7 instance segmentation model was implemented on in-house prepared trees canopy dataset (6500 images) of high resolution remote sensing satellite imagery over Ahmedabad region. During training, evaluation metrics such as bounding box regression and mask regression loss, mean average precision (mAP) and stochastic gradient descent algorithm were used for evaluating and optimizing the performance of model. After the 500 epochs, the mAP of 0.715 and 0.699 for individual tree detection and tree canopy mask segmentation were obtained. There was a challenge to segment individual trees using 50 cm data, which often requires data of 5cm to 15cm resolution. However, by tuning hyper parameters of the model, maximum accuracy of 80 % of trees detection with false segmentation rate of 2% on data was obtained. In future, similar models can be used to obtain better accuracies on better resolution of datasets.

**Funding declaration:** There was no external funding involved in this work.